\documentclass[sn-basic]{sn-jnl}

\jyear{2023}%

\theoremstyle{thmstyleone}%
\theoremstyle{thmstyletwo}%

\theoremstyle{thmstylethree}%

\begin{document}

\title[Towards Phytoplankton Parasite Detection Using Autoencoders]{Towards Phytoplankton Parasite Detection Using Autoencoders}


\author*[1,2]{\fnm{Simon} \sur{Bilik}}\email{bilik@vut.cz}
\equalcont{These authors contributed equally to this work.}

\author[1]{\fnm{Daniel} \sur{Batrakhanov}}\email{daniel.batrakhanov@lut.fi}
\equalcont{These authors contributed equally to this work.}

\author[1]{\fnm{Tuomas} \sur{Eerola}}\email{tuomas.eerola@lut.fi}

\author[3]{\fnm{Lumi} \sur{Haraguchi}}\email{lumi.haraguchi@syke.fi}

\author[3]{\fnm{Kaisa} \sur{Kraft}}\email{kaisa.kraft@syke.fi}

\author[4]{\fnm{Silke} \sur{Van den Wyngaert }}\email{silke.vandenwyngaert@utu.fi}

\author[4]{\fnm{Jonna} \sur{Kangas}}\email{jonna.e.kangas@utu.fi}

\author[4]{\fnm{Conny} \sur{Sjöqvist}}\email{conny.sjoqvist@abo.fi}

\author[5]{\fnm{Karin} \sur{Madsen}}\email{karin.madsen@abo.fi}

\author[1]{\fnm{Lasse} \sur{Lensu}}\email{lasse.lensu@lut.fi}

\author[1]{\fnm{Heikki} \sur{Kälviäinen}}\email{heikki.kalviainen@lut.fi}

\author[2]{\fnm{Karel} \sur{Horak}}
\email{horak@vut.cz}

\affil*[1]{\orgdiv{Computer Vision and Pattern Recognition Laboratory, Department of Computational Engineering}, \orgname{Lappeenranta-Lahti University of Technology LUT}, \orgaddress{\street{Yliopistonkatu 34}, 
\postcode{53850}  
\city{Lappeenranta}, 
\country{Finland}}}

\affil[2]{\orgdiv{Department of Control and Instrumentation, Faculty of Electrical Engineering and Communication}, \orgname{Brno University of Technology}, \orgaddress{\street{Technická 3058/10}, \city{Brno}, \postcode{61600}, \country{Czech Republic}}}

\affil[3]{\orgdiv{Marine Ecology Measurements}, \orgname{Finnish Environment Institute}, \orgaddress{\street{Agnes Sjöbergin Katu 2}, \city{Helsinki}, \postcode{00790}, \country{Finland}}}

\affil[4]{\orgdiv{Department of Biology}, \orgname{University of Turku}, \orgaddress{\street{Vesilinnantie 5}, \city{Turku}, \postcode{20014}, \country{Finland}}}

\affil[5]{\orgdiv{Environmental and Marine Biology
}, \orgname{Åbo Akademi University}, \orgaddress{\street{Henrikinkatu 2}, \city{Turku}, \postcode{20014}, \country{Finland}}}


\abstract{Phytoplankton parasites are largely understudied microbial components with a potentially significant ecological impact on phytoplankton bloom dynamics. To better understand their impact, we need improved detection methods to integrate phytoplankton parasite interactions in monitoring aquatic ecosystems.

Automated imaging devices usually produce high amount of phytoplankton image data, while the occurrence of anomalous phytoplankton data is rare. Thus, we propose an unsupervised anomaly detection system based on the similarity of the original and autoencoder-reconstructed samples. With this approach, we were able to reach an overall F1 score of 0.75 in nine phytoplankton species, which could be further improved by species-specific fine-tuning.

The proposed unsupervised approach was further compared with the supervised Faster R-CNN based object detector. With this supervised approach and the model trained on plankton species and anomalies, we were able to reach the highest F1 score of 0.86. However, the unsupervised approach is expected to be more universal as it can detect also unknown anomalies and it does not require any annotated anomalous data that may not be always available in sufficient quantities.

Although other studies have dealt with plankton anomaly detection in terms of non-plankton particles, or air bubble detection, our paper is according to our best knowledge the first one which focuses on automated anomaly detection considering putative phytoplankton parasites or infections.
}




\maketitle

\section{Introduction}\label{sec1}

Phytoplankton are key players in aquatic systems, mediating biogeochemical cycles and forming the base of the food webs~\citep{falkowski1998biogeochemical}. Phytoplankton population dynamics result from the interplay between resource availability and mortality losses~\citep{reynolds2006ecology}. While some loss mechanisms, such as grazing, are well known, the contribution of other loss mechanisms such as parasitism remains poorly considered and largely understudied in many aquatic systems. Phytoplankton are susceptible to a wide variety of parasites such as viruses, bacteria, protists and fungi. They can cause mortality of certain phytoplankton species, thereby altering phytoplankton bloom dynamics and changing the cycling and flow of energy and matter in aquatic ecosystems~\citep{suttle1990infection},~\citep{klawonn2021characterizing},~\citep{klawonn2023fungal}.

Zoosporic or nanoflagellate parasites infecting phytoplankton encompass a highly diverse functional group of eukaryotic protist and fungal species \citep{scholz2016zoosporic}. They have in common the production of free-living motile stages as their infective propagules which attach to a phytoplankton host cell and further develop either inside (endobiotic) or outside (epibiotic) the host cell, using the host resources for their growth and reproduction. Due to their inconspicuous nature they are difficult to identify and something we are not able to identify typically tends to get overlooked or discarded. Consequently, although the presence and potential importance of these phytoplankton parasites is increasingly recognized, quantitative data in nature is extremely scarce.

An additional challenge includes capturing rapid infection dynamics on a relevant temporal (eg., days) and spatial scale. Obtaining quantitative information of parasite infections by traditional methods is labour-intensive and time-consuming, limiting the spatial and/or temporal coverage of studies investigating phytoplankton-parasite interactions~\citep{peacock2014parasitic}.

Recent technological advances in imaging instruments have made it possible to collect large volumes of plankton image data to study plankton populations, opening new research possibilities \citep{lombard2019globally}. The possibility of high frequency sampling enabled with imaging instruments can allow to better understand phytoplankton dynamics and their potential interaction with parasites~\citep{peacock2014parasitic}. However, while methods for automatic recognition of phytoplankton classes have been widely developed, methods to automatic recognition of parasites infections are still underdeveloped for phytoplankton. This is likely associated to challenges to obtain a sufficient amount of image data of plankton parasites, which requires screening of a huge amount of raw image data, urging for automated solutions for the task.

The scarcity of plankton parasite images is the major challenge for developing deep learning-based computer vision methods for detecting them. While object detection methods such as Faster R-CNN~\citep{ren2015faster} and YOLO~\citep{Jocher_YOLOv5_by_Ultralytics_2020} have been shown to achieve high accuracy on various detection tasks, including parasite detection (see, e.g.,~\citep{bilik2021visual}), they struggle when the amount of training data is limited. Therefore, a more promising approach is to formulate parasite detection as an anomaly detection task. Here the idea is to train the model with images of healthy plankton and to detect images that deviate from the data the models were trained on. Due to the availability of large amounts of plankton image data without parasites for training and relatively small intra-class variation among healthy samples, the images that deviate notably from the training data can be expected to contain potential parasites.

\begin{figure*}[htb]
\centering
\includegraphics[width=1\textwidth]{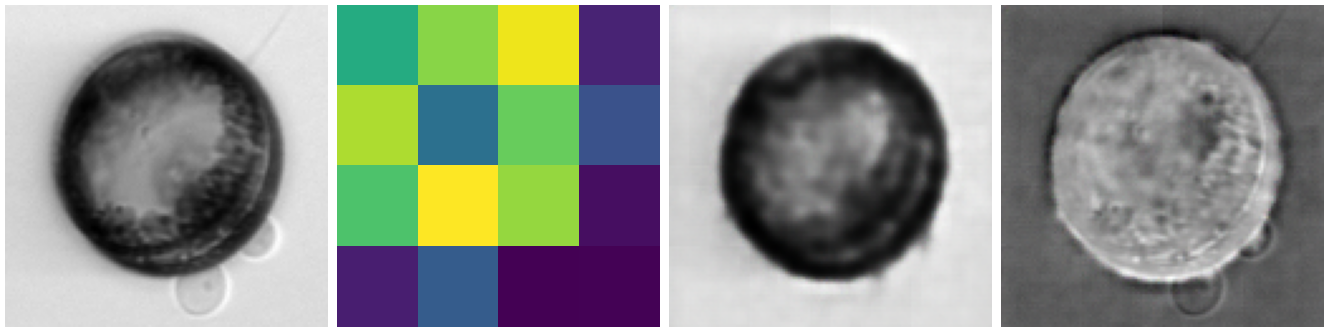}
(a)\hspace{2.5cm}(b)\hspace{2.2cm}(c)\hspace{2.5cm}(d)
\caption{Original (a), encoded space (b), reconstruction (c) and the difference image (d) of the \textit{Centrales} plankton species's anomalous sample.}\label{recCentr}
\end{figure*}

In this work, phytoplankton parasite detection is considered. The problem is formulated as an anomaly detection problem and solved using an autoencoder. The proposed method consists of a vector-quantized variational autoencoder (VQVAE)~\citep{van2017neural} that encodes the input image into a compressed latent representation and uses it to reconstruct the original image. The rationale is that when the autoencoder is trained only on images of healthy phytoplankton it fails to reconstruct the parasites which allows to detect them from the difference image (see Fig.~\ref{recCentr}). The proposed method further employs the HardNet~\citep{mishchuk2017working} feature extractor and Local Outlier Factor~\citep{breunig2000lof} for classifying between healthy plankton and plankton with parasites.

In the experimental part of the work, an extensive set of different backbone convolutional neural networks (CNNs), autoencoder architectures, feature extractors, and classifiers are systematically evaluated on challenging phytoplankton image data to find the best combination to demonstrate the performance of the proposed method. Furthermore, we compare the autoencoder-based anomaly detection method to a Faster R-CNN based object detector. The results show that the proposed method obtains comparable accuracy to the state-of-the-art Faster R-CNN object detector while requiring no images with parasites for training. This makes the autoencoder based method a promising approach for wider utilization in plankton image analysis where collecting large training data of plankton with parasites is infeasible.

\section{Related work}\label{sec2}

Anomaly detection is a data classification technique where a detector models the representation of samples within a specification (OK) and where it distinguishes all samples different from the specification as anomalous (NOK). This problem could be challenging because of potentially high diversity among the NOK samples, imbalance between the number of samples in OK and NOK, and irregularity of the NOK class. A comprehensive overview describing the anomaly detection problems, techniques, and categorization is presented in~\citep{pang2021deep}.

First time introduced on image data in~\citep{hinton2006reducing}, autoencoder (AE) models are widely used in computer vision. The authors of \citep{sakurada2014anomaly} were the first to make the use of worse generalisation ability on different kind out-of-training data of the AE models to detect anomalies in both synthetic and real-world telemetry data using a fully connected AE trained only on the data without anomalies. Based on the results, such an AE can be used to detect previously unseen anomalous samples. This concept was further enhanced and used also on image data, for example, in~\citep{an2015variational} and~\citep{bergmann2018improving}. A comprehensive overview of the AE techniques can be found in~\citep{charte2018practical}.

Plankton anomaly detection has been studied in the context of open-set recognition, i.e., image classification with the present of previously unseen classes (plankton species). In~\citep{pastore2020annotation}, the authors presented an unsupervised approach to classify a plankton sample and to detect potential significant differences (i.e., anomalies) with the respect to the detected class. Image features are extracted using classical computer vision methods and they consists of geometrical, moment-based, and other traditional features. 

In~\citep{pu2021anomaly}, a CNN, trained on the OK samples and the artificial NOK samples derived from the OK data by common data augmentation techniques as blurring and noise adding, is used as the feature extractor. An anomaly score is then computed from those features and it is used together with the trained feature extractor to distinguish between the plankton samples and anomalies. Here air bubbles and non-plankton water particles are considered as anomalies.

In~\citep{pastore2022anomaly}, the authors used a parallel network of custom statistical classifiers called TailDeTect (TDT) to discover previously unseen plankton species. Each of the TDT classifiers is trained on one particular species, and a sample is considered as unknown if none of the classifiers is able to detect it. Unknown samples are collected and validated by experts. Feature extraction and the concept is based on~\citep{pastore2020annotation}. 

In~\citep{badreldeen2022open}, open-set recognition plankton recognition was addressed using a similarity learning approach. Metric learning with angular margin loss was applied to obtain image embedding vectors that model the similarity between images. The anomalies (images from previously unseen classes) were detected by setting a threshold values for the similarity.

Faster R-CNN~\citep{ren2015faster} is a popular deep learning algorithm that has been successfully applied to various domains and tasks, including object detection and anomaly detection. Anomaly detection using Faster R-CNN involves training the model on abnormal images to learn the features of abnormal instances. Then during classification, the model is used to detect abnormal samples that deviate from the expected one. For instance, in industrial manufacturing, abnormal behavior can include machine malfunctions, while in medical diagnosis, it can be unusual patterns in medical images.

For example, in~\citep{zhao2021new}, an improved Faster R-CNN to detect defects in steel plates was used. The algorithm was trained on a dataset of abnormal regions on steel plate images and was able to accurately detect anomalies such as cracks and holes in test images. Similarly, in~\citep{SuYing2021LNDb}, a subtle modification of Faster R-CNN to detect anomalies in CT images of lungs was considered. 

Also object detection methods have been successfully used on the parasite detection, for example, in~\citep{bilik2021visual}. In this case, a YOLOv5 object detector was used to detect a parasitic mite on the honey bee's body. An overview of the other object detection techniques and commonly used dataset could be found for example in~\citep{horak2019deep}.

In plankton research, Faster R-CNN is widely adopted for segmentation and object detection. In ~\citep{li2019developing} several object detection approaches including Faster R-CNN were utilized to evaluate a synthetically augmented dataset. Similar work was presented in~\citep{chen2021object} where a plankton dataset from a dark-field microscope was compiled, and then tested with various object detection methods, including YOLOv3~\citep{redmon2018yolov3}, R-CNN~\citep{girshick2015fast}, and SSD~\citep{liu2016ssd}.

\section{Proposed methods for phytoplankton anomaly detection}\label{sec3}

To detect phytoplankton samples with anomalies, we primarily study an unsupervised autoencoder based approach, followed by different feature extractors and one-class classifiers. To compare the results of our proposed method with a state-of-the-art approach, we utilize supervised object detection based on the Faster R-CNN~\citep{ren2015faster}.

\subsection{Autoencoder-based approach}\label{subsec31}

\begin{figure*}[htb]%
\centering
\includegraphics[width=1\textwidth]{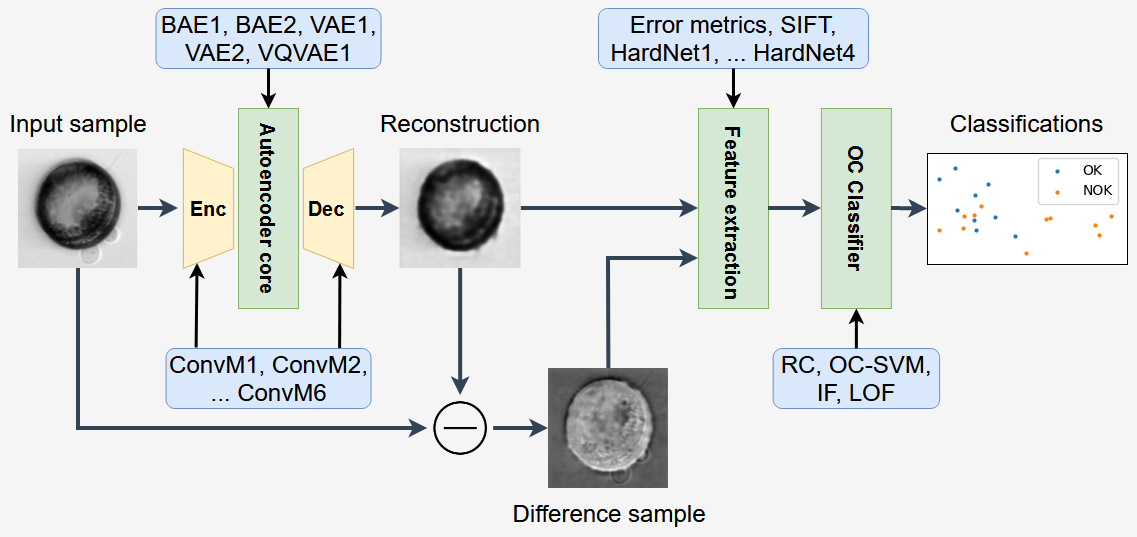}
\caption{The proposed autoencoder-based anomaly detection pipeline.}\label{figFWSch}
\end{figure*}

The proposed method to detect anomalous plankton samples is based on top of the framework available in~\citep{gitAE}. This implementation allows to test various combinations of AE cores, convolutional layers, feature extractors, and one-class classifiers. For the approach, we combined five AE cores, six convolutional encoders and decoders, six feature extractors, and four classifiers (720 combinations in total). The processing pipeline is shown in Fig.~\ref{figFWSch} and it is described in more detail in the sections below.

In the approach, anomaly detection is based on the comparison between the original and autoencoder-reconstructed data, followed by feature extraction and one-class classification.

\subsubsection{Autoencoder architectures and convolutional layers}\label{subsubsec311}

As the first step of anomaly detection, we use AE models trained only on the OK data to reconstruct unknown input samples of both OK and NOK classes. Because of the non-optimal generalisation ability of the AE models and training only on the OK class of data, we hypothesize that data from the NOK class will be reconstructed worse than the data from the OK class as described in~\citep{sakurada2014anomaly}.

To better understand the effect of the AE architecture's core and the complexity of the convolutional encoding and decoding layers, we decided to build our implementation so that the core of the model could be combined with the selected convolutional pairs of the encoders and the decoders. This allows us to analyze contributions of the selected architecture and the convolutional layers separately.

\begin{figure*}[htb]%
\centering
\includegraphics[width=1\textwidth]{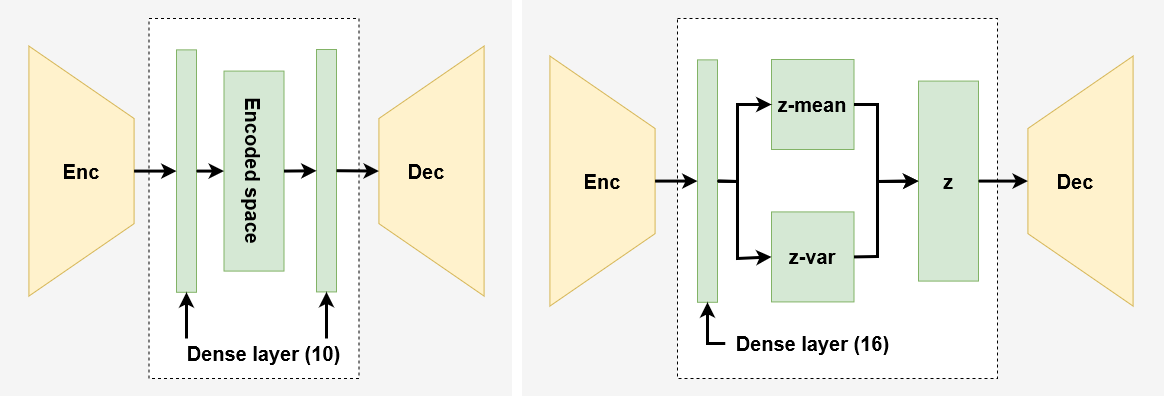}
(a)\hspace{5.5cm}(b)
\caption{Schemes of the modified autoencoder cores: (a) BAE1 core, (b) VAE2 core.}\label{figAESch}
\end{figure*}

For the AE cores, we evaluated five different options. As the first ones, we used implementations of the basic convolutional AE~\citep{masci2011stacked} as the BAE1 core, convolutional variational AE~\citep{pu2016variational} as the VAE1 core and the vector-quantized AE~\citep{van2017neural} as the VQVAE1 core. Besides those cores, we tried to further reduce the features extracted by an encoder by inserting fully-connected layers to the basic convolutional AE as the BAE2 core~\citep{BUT171163} and to the variational AE as the VAE2 core. Described modifications are shown in Fig.~\ref{figAESch}.

We expect that the basic convolutional AE is going to be surpassed by both variational and the vector-quantized cores, because of their non-probabilistic encoding space, which allows the encoding of more anomalies residuum. The quality of reconstructed images should be better in the case of the basic and vector-quantized cores than in the case of variational ones, which typically produce blurry outputs. When training on different classes, the best results are expected from the vector-quantized core, which are supposed to create separable clusters for each class in the encoded space.


Besides the AEs cores described above, we consider six pairs of convolutional encoding and decoding layer architectures, whose structure is described in the complementary tables Table~\ref{tabEnc} for encoders and in Table~\ref{tabDec} for decoders. Each convolutional layer or block described in those tables is complemented with the batch normalization layer. Activation function was set as Leaky ReLu by the ConvM1 architecture and as ReLu for the rest.

The tested convolutional layers go from the most complex ConvM2 suggested for the anomaly detection in~\citep{bergmann2018improving} and ConvM1 architectures, where we expect the ability to reconstruct fine features and details, to the more simple architectures ConvM5, ConvM4 and ConvM3. By the more simple architectures, we expect that the fine features and smaller image structures are going to be suppressed and that they might perform better on the shape or structure anomalies. The last architecture ConvM6 is unsymmetrical as suggested in~\citep{makhzani2015winner} and it uses a more complex encoder of the ConvM5 architecture and a simpler decoder of the ConvM4 architecture. By this architecture, we expect that the anomalies, which would be propagated to the encoded space, will be further suppressed by the decoder reconstruction.


In the optimal case, anomalous areas of the original image are removed during the image reconstruction as shown in Fig.~\ref{recCentr}. A difference image between the original sample and the reconstructed sample is computed and used in the feature extraction.

\subsubsection{Feature extraction}\label{subsubsec312}
The second step the framework applies feature extractors to analyze the reconstructions. The features are based on the comparison between the original and reconstructed data (Error metrics, HardNet3 and HardNet4), or the difference image (SIFT feature extraction, HardNet1 and HardNet2).

The first feature extraction approach (Error metrics) creates a low-dimensional feature vector for each image by computing selected error metrics between the original and reconstructed image. The L2 and SSIM metrics applied in~\citep{BUT171163} are complemented with the Average hash and Mean squared error metrics.

The second feature extraction method (SIFT feature extraction) uses scale and metrics properties of the  image keypoints found by the SIFT method and it is a direct re-implementation presented in~\citep{BUT177722}. This method uses difference images between the original and reconstructed data.

The last four feature extraction methods (HardNet1, HardNet2, HardNet3 and HardNet4) are all based on the batch similarity metric presented in~\citep{mishchuk2017working}. HardNet1 is the most simple one where each sample is described by the HardNet (HN) feature vector of the original image resized to the size of 32x32 as required by the original HN implementation. Since such resizing might not be optimal for small anomalies, the HardNet2 splits the image of the original size to the blocks of 32x32 and computes the HN feature vector for each such block. The resulting feature vector consists of the norms over those vectors. HardNet3 splits the original and reconstructed images to the 32x32 blocks as HardNet2 method, but the resulting feature vector is computed as a cosine similarity between the HN feature vectors of the corresponding blocks of the original and decoded images. HardNet4 uses the same technique, but the cosine similarity is supplemented by the logarithm which is supposed to emphasise smaller differences of the HardNet3 feature vector.

\begin{figure*}[htb]%
\centering
\includegraphics[width=0.8\textwidth]{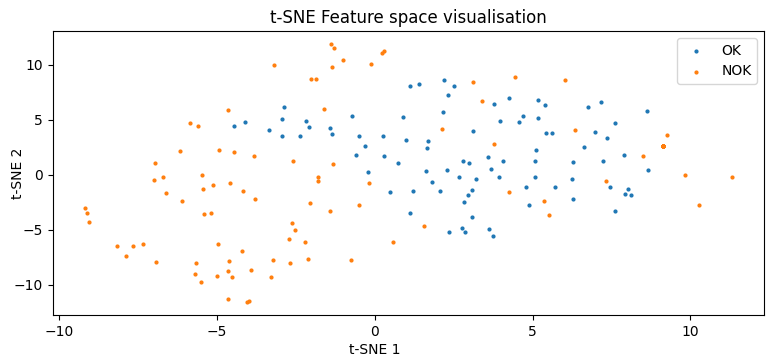}
\caption{Example feature space of the \textit{Aphanizomenon} plankton species.}\label{fsExample}
\end{figure*}

A 2D visualisation of the resulting feature space obtained by the ConvM5-BAE2 autoencoder over the \textit{Aphanizomenon} plankton species using the HardNet2 feature extractor is shown in Fig.~\ref{fsExample}. The OK samples form an elliptical cluster and most of the NOK samples are separated from that cluster.

\subsubsection{One-class classification}\label{subsubsec313}
For the classification part, we used the following one-class classifiers:

\begin{itemize}
\item Robust covariance (RC)~\citep{rousseeuw1999fast}: The RC classifier assumes the same distribution for all OK samples and fits an elliptic envelope to the central data point. The anomaly score is computed using the distribution estimations and Mahalanobis distance.
\item One-class SVM (OC-SVM)~\citep{scholkopf2001estimating}: The OC-SVM classifier utilizes the support vector machine (SVM) and a non-linear kernel to create a separating hyperplane of the training data from the origin of the feature space. Samples on the other side of this hyperplane are considered as anomalies.
\item Isolation Forest (IF)~\citep{liu2008isolation}: The IF classifier uses random feature selection and splitting for isolating observed samples. The anomaly score is based on the total number of splits. Anomalies are supposed to have a smaller number of splits as it should be easier to separate them.
\item Local Outlier Factor (LOF)~\citep{breunig2000lof}: The LOF classifier is based on the local density deviation of the observed point with respect to its k-nearest neighbours. Density of the anomalies should be lower in comparison with the OK samples, which are considered to create denser clusters.
\end{itemize}

The fraction of anomaly samples for the OC-SVM, IF and LOF was set to 1\% being the minimum value of common implementations. We should also assume, that even some OK samples might differ from the majority. All classifiers are fit on the dataset containing only OK samples.

Input features for the one-class classification are normalized using the robust scaling, which normalizes the median and the interquartile range as suggested in~\citep{iglewicz1983robust}. This normalization should be more robust to the outliers than the simple normalization approaches such as the min-max normalization or standardization.

\begin{figure}[htb]%
\centering
\includegraphics[width=0.4\textwidth]{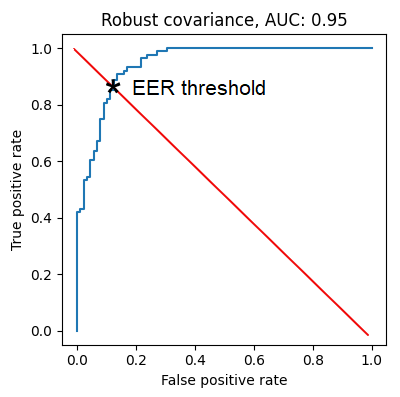}
\caption{Illustration of equal-error-rate (EER) threshold selection criterion on the ROC curve.}\label{figEER}
\end{figure}

In order to select the optimal decision threshold for anomaly detection, we use the Equal error rate (EER) over the ROC curve of the classifier as it is shown in Fig.~\ref{figEER}. All classifiers are fit only on the OK data and the ROC curve was obtained from the test dataset.

\subsection{Object detection based approach}\label{subsec32}

\begin{figure*}[htb]%
\centering
\includegraphics[width=1\textwidth]{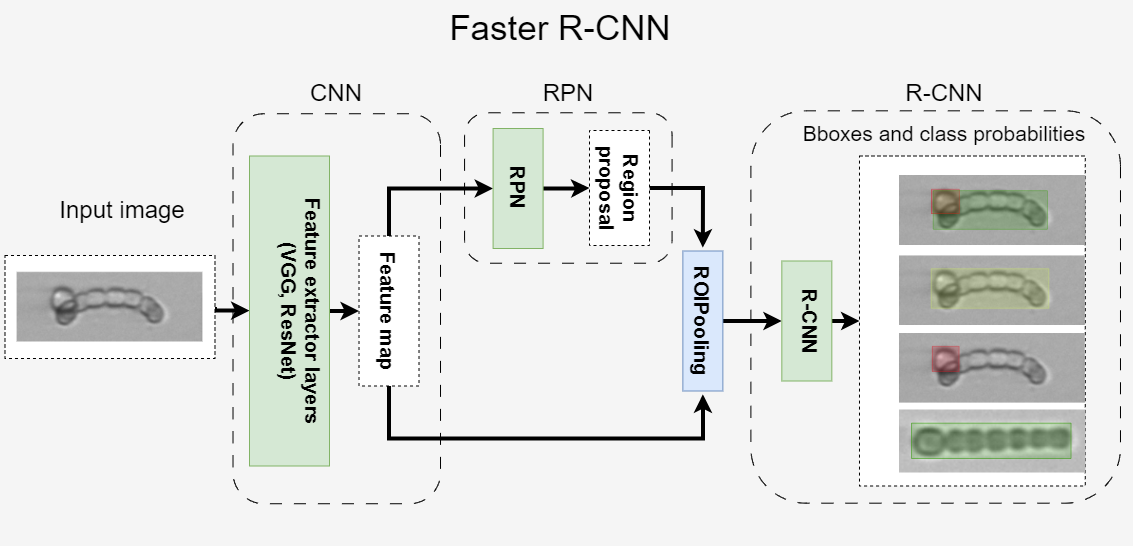}
\caption{Faster R-CNN architecture.}\label{fig:Faster R-CNN}
\end{figure*}

The Faster R-CNN~\citep{ren2015faster} algorithm is composed of three main components: a base feature extractor network, a region proposal network (RPN) for extracting the regions of interest, and a detector that uses the region proposals and respective feature maps to classify the detected objects as shown in Fig.~\ref{fig:Faster R-CNN}. The first component is the feature extractor responsible for generating feature maps from the input image. Usually, this module is a CNN such as VGG-16 or ResNet-50.

The RPN is a kind of fully convolutional network that takes the feature maps from the previous step and returns a set of region proposals that guides the detector on where to find the objects in the image. Then the proposals and corresponding feature maps from the CNN are utilized to yield candidate objects with bounding boxes and fixed-length feature vectors using the ROI Pooling layer. Finally, those outputs are passed to the R-CNN network. The R-CNN network uses the proposed feature maps to classify each bounding box as an object or background and predict final class scores with the bounding boxes.

\begin{figure*}[htb]%
\centering
\includegraphics[width=1\textwidth]{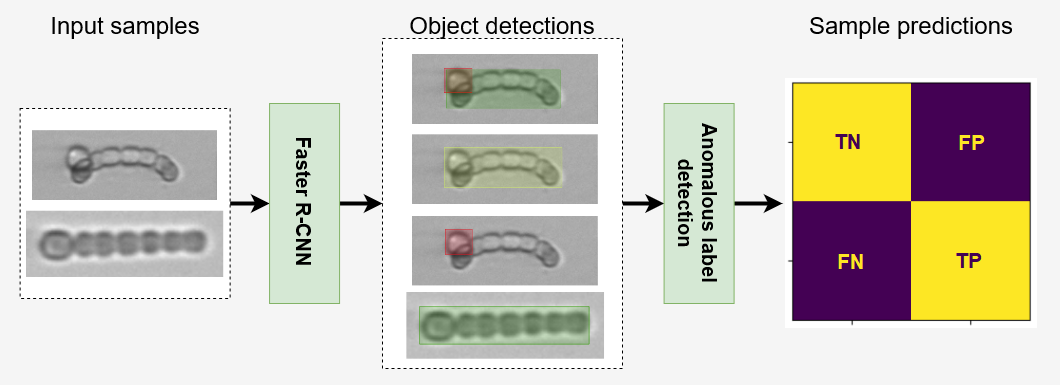}
\caption{The Faster R-CNN approach for anomaly detection.}\label{figRCNNSch}
\end{figure*}

For our object detection experiments, we used the Faster R-CNN implementation available from~\citep{FasterRCNN_PT} based on the ResNet-50 backbone presented in~\citep{li2021benchmarking}. In order to employ an anomaly detection task in the Faster R-CNN baseline, the architecture is supplemented by a one-class classification module based on the predicted object labels as shown in Fig.~\ref{figRCNNSch}.

Since the anomalies such as parasites are relatively small compared to the image size, it is important to consider the anchor generator which is a part of the region-proposal network. Anchors define regions of an image, usually of different aspect ratios and sizes, that are used as references to detect objects. The anchor generator creates a set of anchors for each location in a feature map, then for each region of interest, the model predicts which anchor box best encloses the object. The choice of an anchor generator mostly depends on the type of detection task. For example, if we want to detect small objects, then a smaller anchor size should be used; On the other hand, if the task is to detect objects of various sizes, then a range of anchor sizes should be defined~\citep{ren2015faster}. Additionally, the aspect ratios of the anchors should match the aspect ratios of the objects in the image.

\begin{figure*}[htb]%
\centering
\includegraphics[width=1\textwidth]{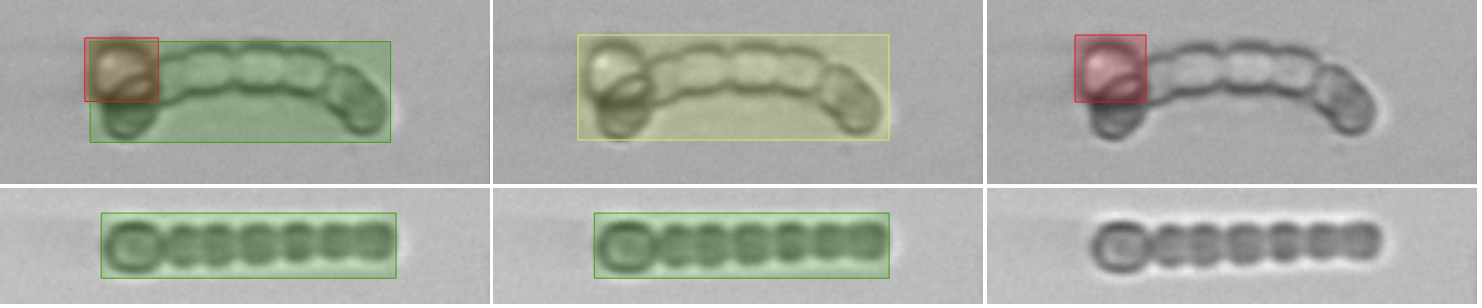}
(a)\hspace{3.5cm}(b)\hspace{3.5cm}(c)
\caption{Object detection tasks using the Faster R-CNN approach:
(a) Plankton vs Anomalies; (b) Plankton vs Anomalous Plankton; (c) Anomalies only.
The NOK samples are shown in the top row and the OK samples in the bottom row.}\label{figRCNNExperiment}
\end{figure*}

As suggested in~\citep{bilik2021visual}, three separate object detectors are considered, each trained on different ground truth: 1) plankton and anomalies, 2) plankton (clean) and anomalous plankton, and 3) anomalies only (see Fig.~\ref{figRCNNExperiment}). In the first column, we can see that the model detects a plankton sample in both cases and an anomaly in the top row. The second column shows a detection of plankton sample with anomaly in the top row and a detection of a clean sample in the bottom row, and finally, the third column shows a detection of a anomaly in the top row only. 

\section{Experiments}\label{sec4}

In this chapter, we describe the used datasets, the evaluation metrics, and the results of the autoencoder based experiments and the object detection based experiments.

\subsection{Phytoplankton anomaly dataset}\label{subsec41}

Natural Baltic Sea phytoplankton communities are continuously imaged with an Imaging FlowCytobot \citep{olson2007submersible} deployed at Ut\"o Atmospheric and Marine Research Station (59°46.84’ N, 21°22.13’ E). The IFCB is connected to the station flow-through system which receives water pumped from a $\sim$5 m deep inlet located 250 m offshore, representative of the sub-surface layer. At Ut\"o, IFCB takes a 5 ml sample nearly every 20 minutes and is set to trigger based on the detection of chlorophyll a, targeting phytoplankton cells rather than non-living particles. The research station and IFCB deployment at Ut\"o are described in detail in~\citep{laakso2018100} and~\citep{kraft2021first}.

The phytoplankton data from Ut\"o IFCB can be currently classified near real-time into 50 different classes as described by~\citep{kraft2022towards}. Putative parasite infection images were manually annotated by experts using another Ut\"o data collected between February--August 2021, using phytoplankton data from nine classes. These classes were selected based on their importance during the spring or summer blooms in the Baltic Sea.

\begin{table*}[htb]
\begin{center}
\begin{minipage}{\textwidth}
\caption{Species-specific statistics of the Plankton anomaly dataset.}\label{tabDS}
\begin{tabular*}{\textwidth}{@{\extracolsep{\fill}}lll@{\extracolsep{\fill}}}
\toprule%
Plankton species & OK samples count & NOK samples count\\
\midrule
Aphanizomenon       & 830                            & 140                             \\
Centrales           & 400                            & 57                              \\
Dolichospermum      & 515                            & 406                             \\
Chaetocero          & 606                            & 371                             \\
Nodularia           & 118                            & 357                             \\
Pauliella           & 160                            & 433                             \\
Peridiniella Chain  & 183                            & 31                              \\
Peridiniella Single & 459                            & 63                              \\
Skeletonem          & 769                            & 419                             \\
\botrule
\end{tabular*}
\end{minipage}
\end{center}
\end{table*}

\begin{figure*}[htb]%
\centering
\includegraphics[width=1\textwidth]{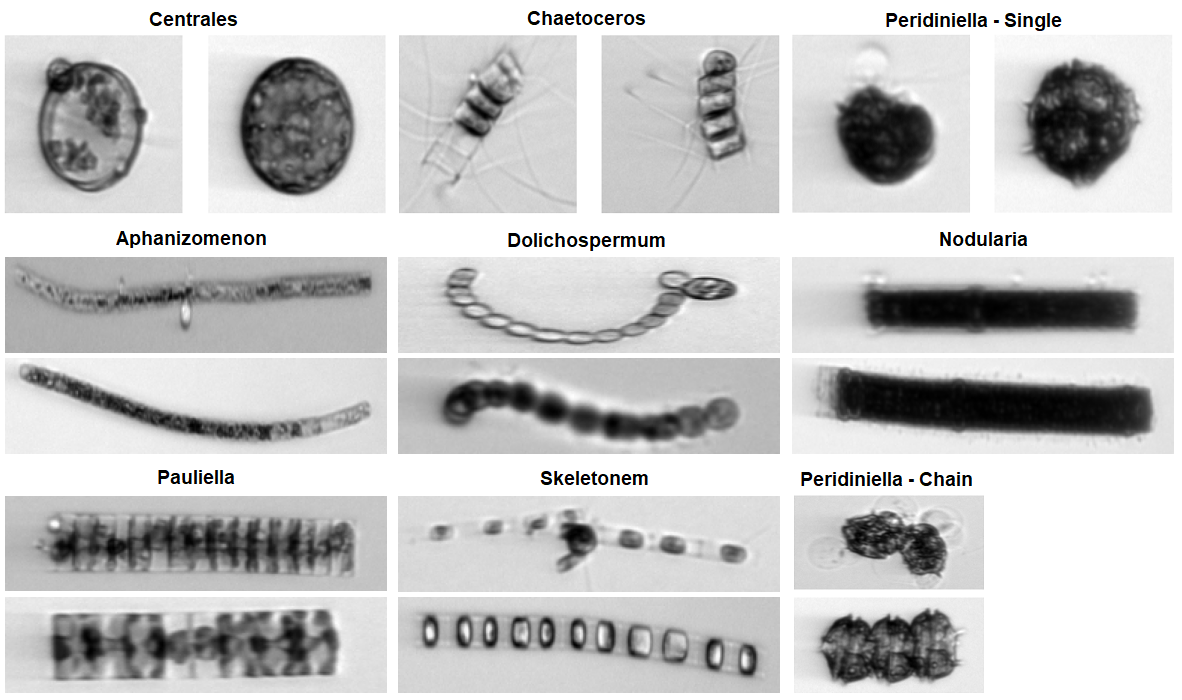}
\caption{Anomalous (left column, or upper row) and non-anomalous samples (right column, or lower row) from all dataset classes of the used dataset.}\label{figDatasetAll}
\end{figure*}

In our experiments, we use a phytoplankton anomaly dataset derived from the annotated images used to train the classifier described in the previous paragraph with the OK samples from the dataset published in~\citep{kraft2022towards} and the NOK samples from the unpublished 2021 Ut\"o data. It contains over 6200 manually annotated and expert-validated samples throughout 9 plankton species with known anomalies as is shown in Table~\ref{tabDS}. Non-anomalous and anomalous samples of each class are shown in Fig.~\ref{figDatasetAll}. As an annotation tool, we used the free version of the Label Studio available at\citep{LabelStudio}. Annotated dataset is available online at~\citep{planktDS} in both COCO and YOLO format.

\subsubsection{Dataset annotations}\label{subsubsec411}

When annotating the dataset we needed to use three kinds of labels to achieve a separate species set as described below:

\begin{itemize}
\item Label \textit{Anomaly} marks the parasite, or other kinds of anomalies on the plankton sample.
\item Label \textit{PlanktonSpecies\_Anomaly} marks the plankton species with the attached parasite.
\item Label \textit{PlanktonSpecies\_Clean} marks the plankton species \textit{with} or \textit{without} the parasite.
\end{itemize}

The last two labels could overlap, but if it was possible, the \textit{PlanktonSpecies\_clean} label does not cover the sample part with parasite. In the need of distinguishing between the OK and NOK sample, \textit{PlanktonSpecies\_clean} label should be removed if it overlaps with the \textit{PlanktonSpecies\_Anomaly} one.

\begin{figure*}[htb]%
\centering
\includegraphics[width=0.8\textwidth]{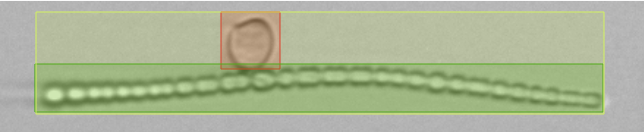}
\caption{Example of the annotation bounding boxes}\label{figAnnotEx}
\end{figure*}

An example of the annotation over a Dolichospermum plankton species sample is shown in Fig.~\ref{figAnnotEx}. The red color marks a plankton anomaly and in this case, the darker green marks the clean sample and the lighter green marks the sample with anomaly.

\subsubsection{Derived dataset for autoencoder-based experiment}\label{subsubsec412}

For the purposes of autoencoder-based experiment, we used the described dataset to derive a one-class dataset with no NOK samples and 70\% of the OK sample in the training set. Test and validation datasets always contain a balanced number of the OK and NOK samples. Experiment with all plankton species contains all available training samples, 10 validation and 10 test samples from each species.

In order to help the AE model to learn more robust features, we added salt-and-pepper noise to the image samples used during the training with a clean sample used as a label as suggested in~\citep{vincent2008extracting}. Besides this noise augmentation, we also use random flipping, contrast, saturation, brightness, inversion, and hue augmentation.

Because the HardNet based feature extractors work correctly only with image sizes of  multiples of 32, all samples were resized with a respect of the major aspect ratio of each class (1:4 by five classes, 1:1 by three classes and 1:2 by one class) as could be seen in Fig.~\ref{figDatasetAll}. For the experiment over all classes, we have chosen the aspect ratio of 1:2 as a compromise.

\subsubsection{Derived dataset for object detection based experiment}\label{subsubsec413}

For the object detection experiment, the model is trained in a supervised manner. The split ratios were set as 70\%, 10\% and 20\% for the training, validation, and test subsets, respectively. Training and validation sets do not include clean samples, while a test contains a balanced number of anomalies and clean images.

Furthermore, we applied the following augmentation techniques: horizontal and vertical flip with a probability 30\%, random brightness, contrast, and saturation adjustment with a probability 10\%.



\subsection{Performance metrics}\label{subsec42}

To compare the results of the autoencoder and object detection experiments, we need to evaluate the predictions of the models with the respect to the ground truth labels. To do so, we can define true positive (TP) and true negative (TN) predictions, where the model correctly classifies OK and NOK samples together with the false positive (FP) and false negative (FN) predictions, where the model misclassifies NOK samples as OK in the FP case and OK samples as NOK in the FN case.


For the comparison between the different variations of autoencoders and object detection methods, the Precision, Recall and F1 score metrics are used. The metrics are defined as follows: 

\begin{equation}
Precision = \frac{\mathit{TP}}{\mathit{TP} + \mathit{FP}} \\
\label{eq:precision}
\end{equation}

\begin{equation}
Recall = \frac{\mathit{TP}}{\mathit{TP} + \mathit{FN}} \\
\label{eq:recall}
\end{equation}

\begin{equation}
F1 = 2 * \frac{Precision * Recall}{Precision + Recall}
\label{eq:F1}
\end{equation}

\bigskip

Similarly as precision and recall, we can also define specificity as follows:
\begin{equation}
Specificity = \frac{\mathit{TN}}{\mathit{TN} + \mathit{FP}} \\
\label{eq:specificity}
\end{equation}

\bigskip

In the autoencoder experiment, we complemented the metrics with the area under curve (AUC) score. This parameter is defined as the area under the Receiver Operator Characteristics (ROC) curve, for which an example is shown in Fig.~\ref{figEER}. This curve is obtained by changing the decision threshold of a binary classifier by a defined step and by plotting the resulting specificity on the x-axis and recall on the y-axis for the each threshold step. Each point of the ROC curve then corresponds to one threshold setting.


\subsection{Autoencoder-based approach}\label{subsec43}

Due to the high number of combinations in our framework (five autoencoder types, six pairs of convolutional layers, six feature extractors and four classifiers), we decided to split the experimental results to two parts. The first part describes the optimal combination of model, feature extractor and classifier trained on all datasets together with its selection criteria whereas the second part describes the optimal results achieved per plankton species. The optimal combination of the autoencoder model, convolutional layers, feature extractor and one-class classifier was determined based on the maximal achieved F1 score. The whole implementation is built using the TensorFlow 2 platform \citep{tensorflow2015-whitepaper} and Scikit-learn library \citep{scikit-learn}.

\subsubsection{Optimal model for anomaly detection} 
\label{subsubsec431}
\begin{figure*}[htb]%
\centering
\includegraphics[width=1\textwidth]{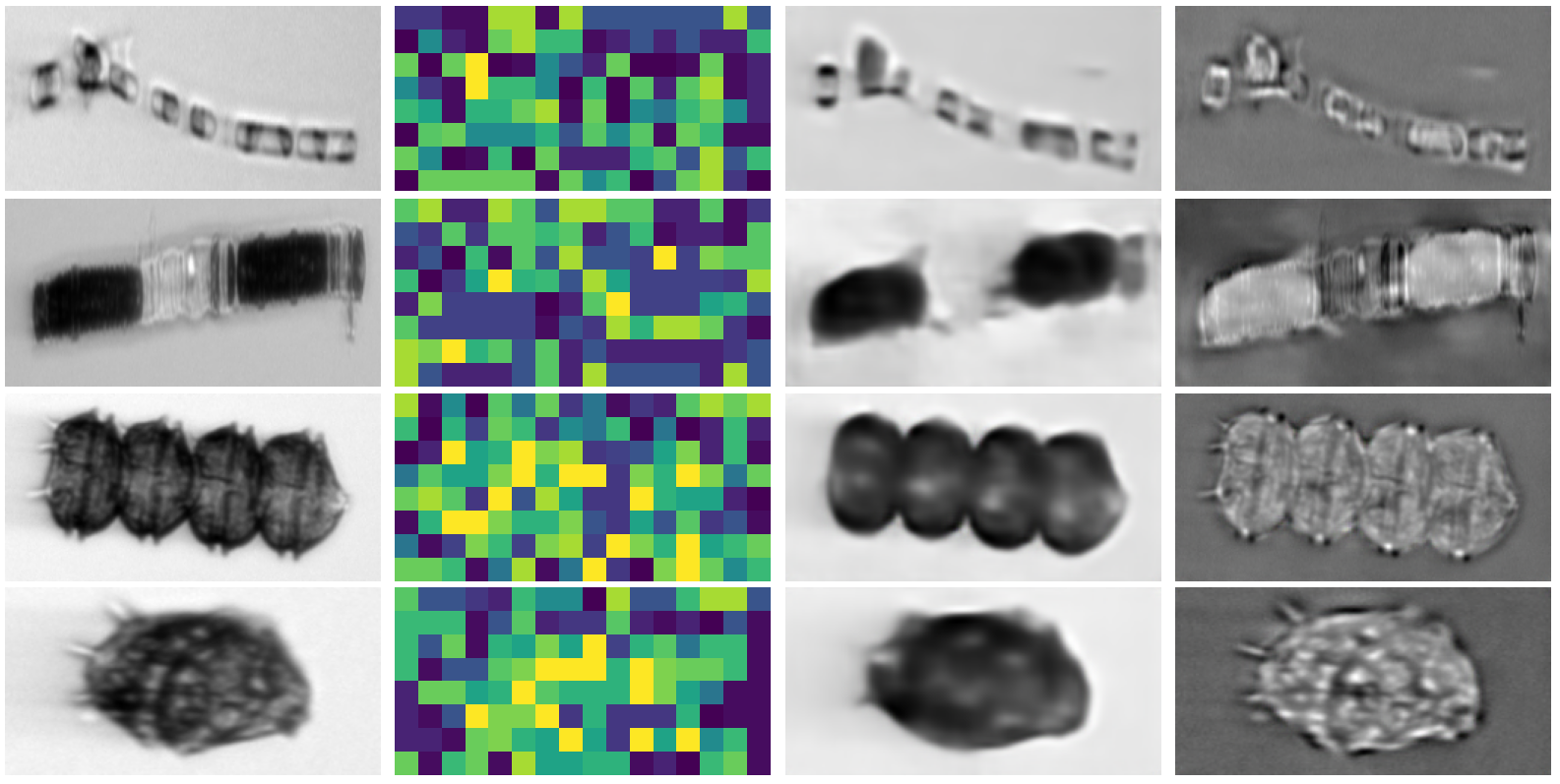}
(a)\hspace{2.5cm}(b)\hspace{2.2cm}(c)\hspace{2.5cm}(d)
\caption{Illustration of the ConvM2-VQVAE1 autoencoder model's results: (a) original, (b) encoded space, (c) reconstruction and (d) difference image.}\label{figModelRes}
\end{figure*}

To select an optimal anomaly detection model, we analyzed the results of all model combinations over the experiment with all plankton species. The best results were achieved with the autoencoder ConvM2-VQVAE1, HardNet1 as a feature extractor and Local Outlier Factor one-class classifier. Those results are shown in Table~\ref{tabR1} and illustrative examples of the results are shown in Fig.~\ref{figModelRes}.

To further demonstrate that the selected anomaly detection model performs best, we evaluated the F1 score over 1) model combinations with a fixed feature extractor and classifier (see Table~\ref{tabR2}), 2) feature extractors combinations with a fixed model and classifier (viz Table~\ref{tabR3}) and 3) classifier combinations with a fixed model and feature extractor (viz Table~\ref{tabR4}). The highest F1 score was 0.75 consistently over all described combinations.

\begin{table*}[htb]
\begin{center}
\begin{minipage}{\textwidth}
\caption{Species specific results with the optimal combination over all plankton species based on the highest F1-score (ConvM2-VQVAE1, HardNet1, Local Outlier Factor).}\label{tabR1}
\begin{tabular*}{\textwidth}{@{\extracolsep{\fill}}lllll@{\extracolsep{\fill}}}
\toprule%
Plankton species       & AUC score & F1 score  & Prec  & Rec   \\ 
\midrule
Aphanizomenon		& 0.89			& 0.83	& 0.83	& 0.83    \\ 
Centrales			& 0.64			& 0.60	& 0.60	& 0.60    \\
Dolichospermum  	& 0.87			& 0.81	& 0.83	& 0.80    \\ 
Chaetocero			& 0.64			& 0.61	& 0.61	& 0.61    \\ 
Nodularia			& 0.70			& 0.67	& 0.67	& 0.67    \\  
Pauliella			& 0.73			& 0.70	& 0.70	& 0.70    \\ 
Peridiniella Chain	& 0.73			& 0.69	& 0.69	& 0.69    \\ 
Peridiniella Single & 0.90			& 0.83	& 0.83	& 0.83    \\ 
Skeletonem			& 0.80			& 0.76	& 0.76	& 0.76    \\ 
\botrule
Plankton all		& 0.80			& 0.75	& 0.75	& 0.74    \\ 
\end{tabular*}
\end{minipage}
\end{center}
\end{table*}

\subsubsection{Optimal anomaly detection model per plankton species}\label{subsubsec432}
Results of the optimal anomaly detection models per plankton species are shown in Table~\ref{tabR5}. The detection results are approximately 10\% better than when using one anomaly detection model trained and all datasets, which could be particularly important for the \textit{Centrales} and \textit{Chaetocero} species with the lowest values of the performance metrics. This is nevertheless traded with a need of a separately trained anomaly detection model for each plankton species.

\subsection{Object detection based approach}\label{subsec44}

Results of the Plankton vs Anomalies, Plankton vs Anomalous Plankton and Anomalies only experiments over all samples are shown in Table~\ref{tabRCNN2}. Plankton vs Anomalies experiment contains both large bounding boxes of plankton annotations and small bounding boxes of anomalies. Therefore, the anchor generator was set up as follows. Sizes of feature map were 16, 32, 64, 128, 256 and 512. In Plankton vs Anomalous Plankton experiment, only the large bounding boxes were used and the sizes were 64, 128, 256, 512, and 1024. For Anomalies only, the sizes were 4, 8, 16, 32, 64, and 128. The scales and the aspect ratios of sizes for each experiment were the same: 0.5, 1.0, 1.5, 2.0, 3.0.

The highest F1 score was achieved with the Plankton vs Anomalies experiment and the lowest one with the Anomalies only experiment. We were able to reach a high F1 score also by Plankton vs Anomalous Plankton experiment, but the resulting plankton labels are often misleading in this case.

\begin{table*}[htb]
\begin{center}
\begin{minipage}{\textwidth}
\caption{Faster R-CNN detection results.}\label{tabRCNN2}
\begin{tabular*}{\textwidth}{@{\extracolsep{\fill}}llllll@{\extracolsep{\fill}}}
\toprule%
Configuration                   & F1 score  & Prec  & Rec   \\ 
\midrule
Plankton vs Anomalies           & \textbf{0.86}	& 0.94	& 0.79    \\
Plankton vs Anomalous Plankton  & 0.83 & 0.82  & 0.85    \\
Anomalies only                  & 0.75	& 0.76	& 0.74    \\
\botrule
\end{tabular*}
\end{minipage}
\end{center}
\end{table*}

\begin{figure}[htb]%
\centering
\includegraphics[width=0.49\textwidth]{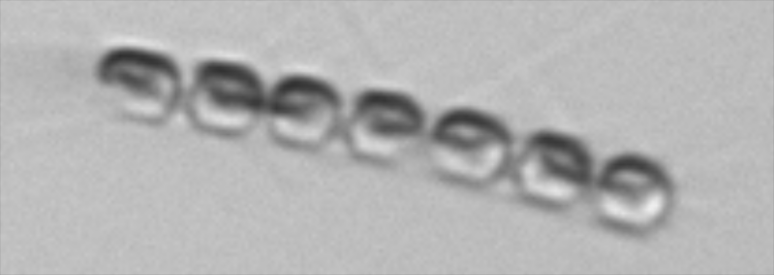}
\includegraphics[width=0.49\textwidth]{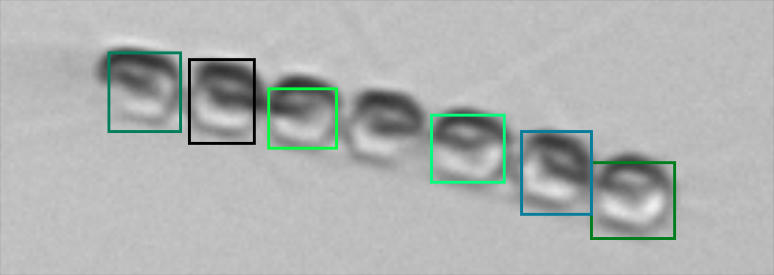}
\includegraphics[width=0.49\textwidth]{fig/confusion_true1.png}
\includegraphics[width=0.49\textwidth]{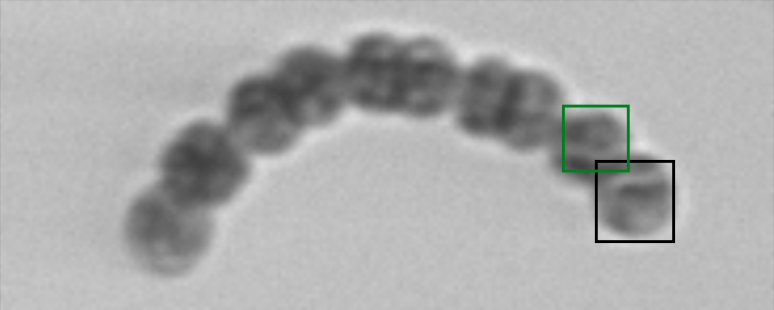}
\caption{Plankton species without any anomalies recognized as an anomaly by Faster R-CNN. The model is confused by a chain of cells and can not distinguish between plankton parts and anomalies.}\label{fig:confusion}
\end{figure}

For the object detection approach, one major issue is that the model is incapable to distinguish between plankton parts and anomalies as shown in Fig.~\ref{fig:confusion}. This major drawback of the Faster R-CNN object detection method originates from the architecture itself and could not be solved by anchor modifications or any other parameter tuning.   

To have a better comparison with Table~\ref{tabR5}, we also performed Anomalies only experiment trained on species-specific data whose results are shown in Table~\ref{tabRCNN1}. In this experiment, the approach performed worse than the species-specific autoencoder experiment and also worse than the universal model within the same experiment on average.


\begin{table*}[htb]
\begin{center}
\begin{minipage}{\textwidth}
\caption{Species-specific dataset Faster R-CNN detection results.}\label{tabRCNN1}
\begin{tabular*}{\textwidth}{@{\extracolsep{\fill}}llllll@{\extracolsep{\fill}}}
\toprule%
Plankton species & F1 score  & Prec  & Rec   \\ 
\midrule
Aphanizomenon		 & 0.66 & 0.63	& 0.68    \\ 
Centrales			 & 0.67	& 0.5	& 1    \\
Dolichospermum  	 & 0.38	& 0.41	& 0.36   \\ 
Chaetocero			 & 0.43	& 0.38	& 0.43   \\ 
Nodularia			 & 0.79 & 0.72	& 0.87    \\  
Pauliella			 & 0.75 & 0.69	& 0.83    \\ 
Peridiniella Chain	 & 0.67	& 0.50	& 1    \\ 
Peridiniella Single  & 0.67 & 0.50	& 1    \\ 
Skeletonem			 & 0.53 & 0.48	& 0.61    \\ 
\midrule
Average              & 0.62 & 0.53  & 0.75    \\
\botrule
\end{tabular*}
\end{minipage}
\end{center}
\end{table*}

We also provide supplementary material as Table~\ref{tab:RCNN-PlanVsAnPlan}, Table~\ref{tab:RCNN-PlanVsAnom} and Table~\ref{tab:RCNN-nok} which shows the results of Plankton vs Anomalies, Plankton vs Anomalous Plankton and Anomalies experiments with the respect to the individual plankton species.

\clearpage

\section{Discussion}\label{sec5}
%
Learning to detect parasites from phytoplankton images is a challenging problem due to large variation in appearance of the plankton cells and parasites, small size of parasites combined with the limited spatial resolution of images, as well as, scarcity of training data caused by the relative rarity of the plankton cells with parasites. 
Even for an expert it is often impossible to confirm the parasitic nature of all the attached (non-host) structures from the images with certainty. For example, spherical structures that are attached to the host cell and have a different appearance to the phytoplankton cell are typically parasites, but can be also loosely attached free living (i.e., non-parasitic) cells or phytoplankton-cell derived organelles expelled from the cell due to stress. This makes it infeasible to collect annotated training and test data just on phytoplankton parasites. Therefore, we formulated the problem as anomaly detection, where the goal is to detect the phytoplankton cells that deviate from "healthy" cells. The method can be seen as an anomaly detector of putative parasite infections allowing to screen large volumes of plankton image data to obtain a subset of interesting images for further analysis.

For the study, a large dataset of phytoplankton cell images with and without anomalies was collected. Untypically for most anomaly detection studies, the collected dataset contains a relatively large amount of images with putative parasites (NOK samples). This made it possible to train also supervised object detectors for the task, and to compare unsupervised anomaly detection methods with the object detector based methods. We evaluated two approaches to detect whether an image contains anomalies or not: 1) autoencoder-based anomaly detection approach and 2) Faster R-CNN-based object detector for anomalies. 

For the autoencoder-based approach, a full pipeline consisting of the CNN-based autoencoder architecture, feature extraction from the reconstruction and difference image, and one-class classifier was proposed. Various methods to each part of the pipeline were considered and extensively evaluated. The best overall accuracy (F1 score) was obtained using the combination of vector-quantized variational autoencoder (VQVAE)~\citep{van2017neural} architecture with the CNN backbone by~\citep{bergmann2018improving}, HardNet feature extractor~\citep{mishchuk2017working} and Local Outlier Factor classifier~\citep{breunig2000lof}. The F1 score for the best combination varied between different species from 0.6 (\textit{Centrales}) to 0.83 (\textit{Peridiniella Single}) with an overall F1 score of 0.75 over all species. The ablation study (Tables~\ref{tabR2-1}-\ref{tabR4}) demonstrated the superiority of the proposed combination over the other alternatives. The accuracy can be further improved by optimizing the method for each phytoplankton species separately as can be seen from Table~\ref{tabR5} with F1 scores varying from 0.73 to 0.94. While fine-tuning class-specific models reduces the generalizability of the method, these are promising results for studying parasites on individual plankton species. 

The limited amount of training data is a notable challenge for the supervised object detectors. To properly learn the large variation in the appearance of anomalies, the training stage would require a sufficient number of example images for each class. The difficulty of the detection task is further emphasized by the fact that the anomalies are typically attached to the host cell and are often very small compared to the plankton cell. These challenges are apparent when observing the Faster R-CNN results (Table~\ref{tabRCNN2}) where the accuracies are not as high as are commonly seen in object detection problems. Three configurations for the R-CNN-based method were evaluated: 1) detection of anomalies only, 2) detection of both anomalies and plankton cells, and 3) detection of healthy plankton cells and plankton cells with anomalies as separate classes. Based on the results it is evident that learning how normal plankton cells look like is beneficial for the R-CNN. The model trained only on anomalies tends to often detect parts of phytoplankton itself as anomalies. It was further noticed that when trained on healthy plankton cells and plankton cells with anomalies as separate classes, the detector often fails to correctly detect the bounding boxes and classify in the presence of anomalies, but still produces correct classification results (OK vs. NOK). This raises questions about the generalizability of the method. 

The Faster R-CNN-based method (model trained to both anomalies and plankton cells) achieved higher accuracy (F1 score: 0.86) than the autoencoder-based method (F1 score: 0.75). This is understandable as the autoencoder-based method did not have access to the images with anomalies during the training stage. Faster R-CNN-based method is more suitable when enough annotated training data is available. This, however, is not typically the case in plankton anomaly and parasite detection due to the reasons discussed above. The autoencoder-based approach has some notable advantages over the supervised object detectors: 1) no training data with anomalies is needed, 2) no annotated bounding boxes are needed, and 3) the method works also with previously unseen anomalies. These together with the comparable accuracy to the R-CNN-based method makes the proposed autoencoder method a more promising approach for plankton anomaly detection on new datasets.

Being able to screen large volumes of plankton image data for anomalies has potential to noticeably reduce the manual work and allows more extensive research on parasitic infections. Ecologically speaking, separating cells with anomalies is interesting and can lead to new research avenues in the future. Anomaly detections can give a valuable first hint of putative parasite infections (or physiologically stressed phytoplankton). However, further method development is needed to make it possible to distinguish between the different types of anomalies and relate them with more certainty to parasites. An interesting future direction would be to apply clustering methods for anomalies to identify different types of anomalies. The classification of anomaly types could be validated by a wider community effort with expertise on different parasite groups, epibionts and symbionts. In combination with classifying different types of anomalies, different datasets could also be collected from known parasites/epibionts/host-derived anomalies from culture systems. Detected parasites and their presence could be further qualitatively confirmed in parallel by additional methods such as microscopy or eDNA-based approaches.


\section{Conclusion}\label{sec6}
This paper presents an unsupervised anomaly detection approach to detect anomalies in nine phytoplankton classes. Although there exists studies on plankton anomaly detection in the context of open-set recognition, i.e. detecting previously unseen plankton classes and non-plankton particles, our paper is according to our best knowledge the first one focusing on the detection of small anomalies such as potential phytoplankton parasites or infections in the known set of plankton classes.  

We propose an anomaly detection pipeline consisting of vector-quantized-variational autoencoder (VQVAE) in combination with the HardNet feature extractor and the Local Outlier Factor classifier. With this pipeline, we achieved an average F1 score of 0.75 for all nine analyzed phytoplankton species. We also suggest that the achieved anomaly detection results could be further improved by optimizing the components of the pipeline for each species separately.

The results achieved with this approach were compared to a supervised Faster R-CNN object detector experimented in three configurations: 1) detection of anomalies only, 2) detection of anomalies and plankton cells, and 3) detection of plankton cells with and without anomalies. The best results were achieved with the second configuration with an average F1 score of 0.86. Although this score is higher than the one achieved by the the suggested autoencoder, our approach is more universal because it can detect also previously unseen data and it needs no anomalous samples for its training. These benefits make the proposed autoencoder approach more promising for plankton research where the annotated anomaly datasets are not available or not feasible to collect. We have made our code and used dataset publicly available as a part of this paper.

\backmatter

\section*{Acknowledgments}
The research was carried out in the FASTVISION and FASTVISION-plus projects funded by the Academy of Finland (Decision numbers 321980, 321991, 339612, and 339355).
The work was further supported by the grant number FEKT-S-23-8451 "Research on advanced methods and technologies in cybernetics, robotics, artificial intelligence, automation and measurement" from the Internal science fund of Brno University of Technology.





\clearpage

\begin{appendices}

\section{Convolutional encoders and decoders}\label{secA1}

\begin{table*}[htb]
\begin{center}
\begin{minipage}{\textwidth}
\caption{Overview of the tested convolutional encoder models}\label{tabEnc}
\begin{tabular*}{\textwidth}{@{\extracolsep{\fill}}lllcc@{\extracolsep{\fill}}}
\toprule%
Encoder name & Layer name & Filters & Kernel size & Stride\\
\midrule
\multirow{5}{*}{ConvM1}  & ConvE1   & 32    & 3x3   & 2 \\
    & ConvE2   & 64    & 3x3   & 2 \\
    & ConvE3   & 64    & 3x3   & 2 \\
    & ConvE4   & 64    & 3x3   & 2 \\
    & ConvE5   & 64    & 3x3   & 2 \\
\hline

\multirow{9}{*}{ConvM2}  & ConvE1   & 32    & 4x4   & 2 \\
    & ConvE2   & 32    & 4x4   & 2 \\
    & ConvE3   & 32    & 3x3   & 1 \\
    & ConvE4   & 64    & 4x4   & 2 \\
    & ConvE5   & 64    & 3x3   & 1 \\
    & ConvE6   & 128   & 4x4   & 2 \\
    & ConvE7   & 64    & 3x3   & 1 \\
    & ConvE8   & 32    & 3x3   & 1 \\
    & ConvE8   & 1     & 8x8   & 1 \\
\hline

\multirow{2}{*}{ConvM3}  & ConvE1   & 32    & 3x3   & 2 \\
    & ConvE2   & 64    & 3x3   & 2 \\
\hline

\multirow{4}{*}{ConvM4}  & ConvE1   & 8    & 5x5   & 1 \\
    & MaxPool  & -     & 2x2   & - \\
    & ConvE2   & 4     & 3x3   & 1 \\
    & MaxPool  & -     & 2x2   & - \\
\hline

\multirow{6}{*}{ConvM5}  & ConvE1   & 16    & 3x3   & 1 \\
    & MaxPool  & -     & 2x2   & - \\
    & ConvE2   & 8     & 3x3   & 1 \\
    & MaxPool  & -     & 2x2   & - \\
    & ConvE3   & 4     & 3x3   & 1 \\
    & MaxPool  & -     & 2x2   & - \\

\botrule
\end{tabular*}
\end{minipage}
\end{center}
\end{table*}

\begin{table*}[htb]
\begin{center}
\begin{minipage}{\textwidth}
\caption{Overview of the tested convolutional decoder models}\label{tabDec}
\begin{tabular*}{\textwidth}{@{\extracolsep{\fill}}lllcc@{\extracolsep{\fill}}}
\toprule%
Decoder name & Layer name & Filters & Kernel size & Stride\\
\midrule
\multirow{5}{*}{ConvM1}  & ConvT1   & 64    & 3x3   & 2 \\
    & ConvT2   & 64    & 3x3   & 2 \\
    & ConvT3   & 64    & 3x3   & 2 \\
    & ConvT4   & 32    & 3x3   & 2 \\
    & ConvT5   & Image channels    & 3x3   & 2 \\
\hline

\multirow{16}{*}{ConvM2}  & ConvD1   & 16    & 3x3   & 1 \\
    & ConvD2        & 64    & 3x3   & 1 \\
    & Upsampling    & -     & 2x2   & - \\
    & ConvD3        & 128   & 4x4   & 1 \\
    & Upsampling    & -     & 2x2   & - \\
    & ConvD4        & 64    & 3x3   & 1 \\
    & Upsampling    & -     & 2x2   & - \\
    & ConvD5        & 64    & 4x4   & 1 \\
    & Upsampling    & -     & 2x2   & - \\
    & ConvD6        & 32    & 3x3   & 1 \\
    & Upsampling    & -     & 2x2   & - \\
    & ConvD7        & 32    & 4x4   & 1 \\
    & Upsampling    & -     & 2x2   & - \\
    & ConvD8        & 32    & 4x4   & 1 \\
    & Upsampling    & -     & 2x2   & - \\
    & ConvD8        & Image channels     & 8x8   & 1 \\
\hline

\multirow{4}{*}{ConvM3}  & ConvD1   & 64    & 3x3   & 1 \\
    & ConvD2        & 32    & 3x3   & 1 \\
    & Upsampling    & -     & 2x2   & - \\
    & ConvD3        & Image channels    & 3x3   & 1 \\
\hline

\multirow{5}{*}{ConvM4}  & ConvD1   & 4    & 3x3   & 1 \\
    & Upsampling    & -     & 2x2   & - \\
    & ConvD2        & 8     & 3x3   & 1 \\
    & Upsampling    & -     & 2x2   & - \\
    & ConvD3        & Image channels    & 3x3   & 1 \\
\hline

\multirow{7}{*}{ConvM5}  & ConvD1   & 4    & 3x3   & 1 \\
    & Upsampling    & -     & 2x2   & - \\
    & ConvD2        & 8     & 3x3   & 1 \\
    & Upsampling    & -     & 2x2   & - \\
    & ConvD3        & 16    & 3x3   & 1 \\
    & Upsampling    & -     & 2x2   & - \\
    & ConvD4        & Image channels    & 3x3   & 1 \\

\botrule
\end{tabular*}
\end{minipage}
\end{center}
\end{table*}

\clearpage

\section{Complementary results of the autoencoder based experiment}\label{secA2}

\begin{table*}[htb]
\begin{center}
\begin{minipage}{\textwidth}
\caption{Results over all plankton species for fixed feature extractor (HardNet1) and classifier (Local Outlier Factor).}\label{tabR2}
\begin{tabular*}{\textwidth}{@{\extracolsep{\fill}}lllll@{\extracolsep{\fill}}}
\toprule%
Model name     & AUC score & F1 score  & Prec  & Rec   \\ 
\midrule
ConvM1-BAE1		& 0.73			& 0.70	& 0.70	& 0.70 \\ 
ConvM1-BAE2		& 0.81			& 0.72	& 0.72	& 0.72 \\ 
ConvM1-VAE1		& 0.74			& 0.66	& 0.66	& 0.66 \\ 
ConvM1-VAE2		& 0.72			& 0.65	& 0.65	& 0.64 \\ 
ConvM1-VQVAE1	& 0.72			& 0.64	& 0.64	& 0.64 \\ 
ConvM2-BAE1		& 0.73			& 0.67	& 0.67	& 0.67 \\ 
ConvM2-BAE2		& 0.79			& 0.72	& 0.72	& 0.72 \\ 
ConvM2-VAE1		& 0.72			& 0.67	& 0.67	& 0.67 \\ 
ConvM2-VAE2		& 0.79			& 0.70	& 0.71	& 0.70 \\ 
ConvM2-VQVAE1	& 0.80			& \textbf{0.75} & 0.75	& 0.74 \\ 
ConvM3-BAE1		& 0.68			& 0.60	& 0.60	& 0.60 \\ 
ConvM3-BAE2		& 0.69			& 0.62	& 0.62	& 0.62 \\ 
ConvM3-VAE1		& 0.72			& 0.67	& 0.67	& 0.67 \\ 
ConvM3-VAE2		& 0.69			& 0.63	& 0.63	& 0.63 \\ 
ConvM3-VQVAE1	& 0.71			& 0.65	& 0.66	& 0.64 \\ 
ConvM4-BAE1		& 0.70			& 0.61	& 0.62	& 0.61 \\ 
ConvM4-BAE2		& 0.72			& 0.64	& 0.65	& 0.63 \\ 
ConvM4-VAE1		& 0.69			& 0.64	& 0.64	& 0.64 \\ 
ConvM4-VAE2		& 0.71			& 0.64	& 0.64	& 0.64 \\ 
ConvM4-VQVAE1	& 0.74			& 0.71	& 0.71	& 0.71 \\ 
ConvM5-BAE1		& 0.74			& 0.69	& 0.69	& 0.69 \\ 
ConvM5-BAE2		& 0.69			& 0.64	& 0.64	& 0.64 \\ 
ConvM5-VAE1		& 0.66			& 0.57	& 0.57	& 0.58 \\ 
ConvM5-VAE2		& 0.74			& 0.68	& 0.68	& 0.68 \\ 
ConvM5-VQVAE1	& 0.72			& 0.68	& 0.69	& 0.68 \\ 
ConvM6-BAE1		& 0.70			& 0.64	& 0.64	& 0.64 \\ 
ConvM6-BAE2		& 0.73			& 0.68	& 0.68	& 0.68 \\ 
ConvM6-VAE1		& 0.72			& 0.69	& 0.69	& 0.69 \\ 
ConvM6-VAE2		& 0.75			& 0.70	& 0.70	& 0.70 \\ 
ConvM6-VQVAE1	& 0.71			& 0.67	& 0.67	& 0.67 \\ 
\botrule
\end{tabular*}
\end{minipage}
\end{center}
\end{table*}

\begin{table*}[htb]
\begin{center}
\begin{minipage}{\textwidth}
\caption{Results over all plankton species for fixed convolutional layer architecture (ConvM2), feature extractor (HardNet1) and classifier (Local Outlier Factor).}\label{tabR2-1}
\begin{tabular*}{\textwidth}{@{\extracolsep{\fill}}lllll@{\extracolsep{\fill}}}
\toprule%
Model architecture     & AUC score & F1 score  & Prec  & Rec   \\ 
\midrule
BAE1		& 0.73			& 0.67	& 0.67	& 0.67 \\ 
BAE2		& 0.79			& 0.72	& 0.72	& 0.72 \\ 
VAE1		& 0.72			& 0.67	& 0.67	& 0.67 \\ 
VAE2		& 0.79			& 0.70	& 0.71	& 0.70 \\ 
VQVAE1	& 0.80			& \textbf{0.75} & 0.75	& 0.74 \\ 
\botrule
\end{tabular*}
\end{minipage}
\end{center}
\end{table*}

\begin{table*}[htb]
\begin{center}
\begin{minipage}{\textwidth}
\caption{Results over all plankton species for fixed model architecture (VQVAE1), feature extractor (HardNet1) and classifier (Local Outlier Factor).}\label{tabR2-2}
\begin{tabular*}{\textwidth}{@{\extracolsep{\fill}}lllll@{\extracolsep{\fill}}}
\toprule%
Conv. architecture     & AUC score & F1 score  & Prec  & Rec   \\ 
\midrule
ConvM1	& 0.72			& 0.64	& 0.64	& 0.64 \\ 
ConvM2	& 0.80			& \textbf{0.75} & 0.75	& 0.74 \\ 
ConvM3	& 0.71			& 0.65	& 0.66	& 0.64 \\ 
ConvM4	& 0.74			& 0.71	& 0.71	& 0.71 \\ 
ConvM5	& 0.72			& 0.68	& 0.69	& 0.68 \\ 
ConvM6	& 0.71			& 0.67	& 0.67	& 0.67 \\ 
\botrule
\end{tabular*}
\end{minipage}
\end{center}
\end{table*}

\begin{table*}[htb]
\begin{center}
\begin{minipage}{\textwidth}
\caption{Results over all plankton species for fixed model (ConvM2-VQVAE1) and classifier (Local Outlier Factor).}\label{tabR3}
\begin{tabular*}{\textwidth}{@{\extracolsep{\fill}}lllll@{\extracolsep{\fill}}}
\toprule%
Feature extractor      & AUC score & F1 score  & Prec  & Rec   \\  
\midrule
ErrMetrics				& 0.50			& 0.51	& 0.51	& 0.51 \\ 
SIFT				    & 0.49			& 0.43	& 0.43	& 0.43 \\ 
HardNet1				& 0.80			& \textbf{0.75}	& 0.75	& 0.74 \\ 
HardNet2				& 0.64			& 0.60	& 0.60	& 0.60 \\ 
HardNet3				& 0.75			& 0.68	& 0.70	& 0.67 \\ 
HardNet4				& 0.77			& 0.73	& 0.73	& 0.72 \\ 
\botrule
\end{tabular*}
\end{minipage}
\end{center}
\end{table*}

\begin{table*}[htb]
\begin{center}
\begin{minipage}{\textwidth}
\caption{Results over all plankton species for fixed model (ConvM2-VQVAE1) and feature extractor (HardNet1).}\label{tabR4}
\begin{tabular*}{\textwidth}{@{\extracolsep{\fill}}lllll@{\extracolsep{\fill}}}
\toprule%
Classifier         & AUC score & F1 score  & Prec  & Rec   \\  
\midrule
Robust covariance		& 0.66		& 0.61	& 0.62	& 0.61    \\ 
One-Class SVM		    & 0.59		& 0.56	& 0.56	& 0.57    \\ 
Isolation Forest		& 0.54		& 0.53	& 0.53	& 0.52    \\ 
Local Outlier Factor	& 0.80		& \textbf{0.75}	& 0.75	& 0.74    \\ 
\botrule
\end{tabular*}
\end{minipage}
\end{center}
\end{table*}

\clearpage

\begin{sidewaystable*}[htb]
\sidewaystablefn%
\begin{center}
\begin{minipage}{\textwidth}
\caption{Overview of the per-plankton-species optimal anomaly detection combinations and their results.}\label{tabR5}
\begin{tabular*}{1.0\textwidth}{@{\extracolsep{\fill}}llllllll@{\extracolsep{\fill}}}
\toprule%
Plankton species         & Model name     & Feature extractor      & Classifier         & AUC score & F1 score  & Prec  & Rec  \\
\midrule
Aphanizomenon		& ConvM5-BAE1			& HardNet2			& LOF	& 0.95			& 0.93	& 0.91	& 0.94  \\
Centrales		    & ConvM2-VAE1			& HardNet2			& RC		& 0.74			& 0.73	& 0.72	& 0.72  \\ 
Dolichospermum		& ConvM6-BAE1			& HardNet1			& LOF	& 0.93			& 0.90	& 0.92	& 0.88  \\ 
Chaetocero		    & ConvM1-BAE1			& HardNet1			& RC		& 0.72			& 0.72	& 0.71	& 0.72  \\ 
Nodularia		    & ConvM3-BAE1			& HardNet1			& RC		& 0.95			& 0.94	& 1.00	& 0.89  \\  
Pauliella		    & ConvM3-VAE1			& HardNet3			& RC		& 0.83			& 0.86	& 0.81	& 0.91  \\ 
Peridiniella Chain	& ConvM1-VAE1			& HardNet4			& IF		& 0.93			& 0.88	& 0.88	& 0.88  \\ 
Peridiniella Single	& ConvM6-BAE1			& HardNet1			& LOF	& 0.90			& 0.86	& 0.84	& 0.89  \\ 
Skeletonem		    & ConvM6-VAE2			& HardNet4			& RC		& 0.90			& 0.84	& 0.84	& 0.84  \\ 
\botrule
\end{tabular*}
\end{minipage}
\end{center}
\end{sidewaystable*}

\clearpage

\section{Complementary R-CNN results}\label{secA4}

\begin{table*}[htb]
\begin{center}
\begin{minipage}{\textwidth}
\caption{Faster R-CNN detection results for Plankton vs Anomalies experiment.}\label{tab:RCNN-PlanVsAnom}
\begin{tabular*}{\textwidth}{@{\extracolsep{\fill}}llllll@{\extracolsep{\fill}}}
\toprule%
Plankton species & F1 score  & Prec  & Rec   \\ 
\midrule
Aphanizomenon		 & 0.98 & 1	& 0.96    \\ 
Centrales			 & 0.85	& 0.73	& 1    \\
Dolichospermum  	 & 0.89	& 0.97	& 0.81   \\ 
Chaetocero			 & 0.80	& 0.89	& 0.73   \\ 
Nodularia			 & 0.76 & 0.98	& 0.62    \\  
Pauliella			 & 0.86 & 0.99	& 0.77    \\ 
Peridiniella Chain	 & 0.60	& 0.75	& 0.5    \\ 
Peridiniella Single  & 0.76 & 1	& 0.62    \\ 
Skeletonem			 & 0.98 & 0.97	& 1    \\ 
\midrule
Plankton \& Anomalies  & 0.87 & 0.95 & 0.81    \\
\botrule
\end{tabular*}
\end{minipage}
\end{center}
\end{table*}

\begin{table*}[htb]
\begin{center}
\begin{minipage}{\textwidth}
\caption{Faster R-CNN detection results for Plankton vs Anomalous Plankton experiment.}\label{tab:RCNN-PlanVsAnPlan}
\begin{tabular*}{\textwidth}{@{\extracolsep{\fill}}llllll@{\extracolsep{\fill}}}
\toprule%
Plankton species & F1 score  & Prec  & Rec   \\ 
\midrule
Aphanizomenon		 & 0.66 & 0.58	& 0.75    \\ 
Centrales			 & 0.60	& 0.47	& 0.82    \\
Dolichospermum  	 & 0.85	& 0.82	& 0.88   \\ 
Chaetocero			 & 0.74	& 0.71	& 0.78   \\ 
Nodularia			 & 0.78 & 0.96	& 0.66    \\  
Pauliella			 & 0.79 & 0.76	& 0.83    \\ 
Peridiniella Chain	 & 0.67	& 0.67	& 0.67    \\ 
Peridiniella Single  & 0.88 & 0.92	& 0.85    \\ 
Skeletonem			 & 0.94 & 0.92	& 0.94    \\ 
\midrule
Plankton \& Anomalous plankton & 0.83 & 0.82 & 0.85    \\ 
\botrule
\end{tabular*}
\end{minipage}
\end{center}
\end{table*}

\begin{table*}[htb]
\begin{center}
\begin{minipage}{\textwidth}
\caption{Faster R-CNN detection results for Anomalies experiment.}\label{tab:RCNN-nok}
\begin{tabular*}{\textwidth}{@{\extracolsep{\fill}}llllll@{\extracolsep{\fill}}}
\toprule%
Plankton species & F1 score  & Prec  & Rec   \\ 
\midrule
Aphanizomenon		 & 1 & 1	& 1    \\ 
Centrales			 & 0.77	& 0.67	& 0.91    \\
Dolichospermum  	 & 0.38	& 0.87	& 0.25   \\ 
Chaetocero			 & 0.67	& 0.66	& 0.68   \\ 
Nodularia			 & 0.85 & 0.88	& 0.82    \\  
Pauliella			 & 0.85 & 0.78	& 0.94    \\ 
Peridiniella Chain	 & 0.44	& 0.67	& 0.33    \\ 
Peridiniella Single  & 0.89 & 0.86 & 0.92    \\ 
Skeletonem			 & 0.83 & 0.75	& 0.93    \\ 
\midrule
Anomalies & 0.75 & 0.76 & 0.74    \\
\botrule
\end{tabular*}
\end{minipage}
\end{center}
\end{table*}




\end{appendices}


\clearpage

\end{document}